# Reflections on Inductive Thematic Saturation as a potential metric for measuring the validity of an inductive Thematic Analysis with LLMs

Stefano De Paoli – Abertay University – s.depaoli@abertay.ac.uk

Walter Stan Mathis – Yale Medical School - walter.mathis@yale.edu

[as this is a draft work-in-progress we are posting to foster discussion, please send us comments, observations or if you spot errors, especially on how we propose to calculate the Probability problem]

**Abstract**

This paper presents a set of reflections on saturation and the use of Large Language Models (LLMs) for performing Thematic Analysis (TA). The paper suggests that initial thematic saturation (ITS) could be used as a metric to assess part of the transactional validity of TA with LLM, focusing on the initial coding. The paper presents the initial coding of two datasets of different sizes, and it reflects on how the LLM reaches some form of analytical saturation during the coding. The procedure proposed in this work leads to the creation of two codebooks, one comprising the total cumulative initial codes and the other the total unique codes. The paper proposes a metric to synthetically measure ITS using a simple mathematical calculation employing the ratio between slopes of cumulative codes and unique codes. The paper contributes to the initial body of work exploring how to perform qualitative analysis with LLMs.

## Introduction

There is a nascent body of work exploring the use of Large Language Models (LLMs) for performing qualitative coding (QC), Thematic Analysis (TA) and a few other types of Qualitative Analysis (QA). LLMs are complex and sophisticated Machine Learning systems trained on a large amount of textual data, can understand natural language, and generate linguistic responses. Some known examples include the OpenAI GPT models (which are at the basis of ChatGPT), Palm (which is at the basis of Google Bard) or Llama-2 from Meta/Facebook.

While initial evidence shows that something satisfactory can be derived by using LLMs for the purpose of doing TA, we still are in a phase of methodological research where we need to consider if the output quality, we receive from LLMs is valid. In other words, we need to assess if an LLM-performed TA is 'good enough': if a QC or TA performed with LLMs can have comparable validity to one produced by an expert human researcher. Comparing themes produced by human researchers and LLMs for TA has been used for example by the authors adopting a qualitative textual comparison (De Paoli, 2023a). Drapal et al. (2023) used an LLM to support their TA of court cases, evaluating also the results against human coder. In this paper we argue that a further potential angle for assessing the quality and validity of LLM-driven QC analysis could be considering the well-established notion of **saturation**.

The widespread understanding of saturation often comes from the definition given in the context of Grounded Theory where this is understood to be "when no new information seems to emerge from coding, that is, when no new properties, dimensions, conditions, actions/interactions or consequences are seen in the data" (Strauss and Corbin, 1990, p. 136). This is very often understood as a metric relating to the sampling of data, namely if saturation is not reached (i.e. new information and consequently codes keep emerging from analysing interviews) researchers are potentially advised to collect further data, since the existing sample of interviews is probably not enough to describe or characterise the phenomenon under study, or to answer the research question. In practice, however, there are various constraints to the application of the saturation on the sample, such as the fact that often the number of e.g. interviews to be conducted in a defined piece of research is declared in advance (Charmaz, 2014) for example to funding agencies. Braun and Clarke (2021) also argued that data saturation is not always a significant metric for validity for TA.

Empirical studies also have been conducted to assess how many interviews would be enough to reach saturation to identify the optimal number of interviews. However, a clear agreement across the research community does not seem to exist. For example, Hagaman et al. (2017), established that 16 interviews were sufficient (in their research) to reach saturation and find common themes. A recent systematic review of literature by Hennik and Kaiser (2022) identified that across all the reviewed readings "an average of 12–13 interviews reached saturation". Also, mathematical/statistical models have been defined to measure saturation (see e.g. Guest et al., 2020 – which in addition offer a good overview of the literature on determining saturation). A useful discussion is also offered by Saunders et al. (2018) which make a distinction between four types of saturation: (1) theoretical saturation, which we can associate with measuring the **sampling**; (2) inductive thematic saturation which relates to the emergence of new codes/categories and therefore has an **analytical** purpose**;** (3) a priori thematic saturation, which relates to how the data exemplify a pre-defined set of coding categories, and thus also has a **sampling** nature and (4) data saturation which represents a measure of data redundancy and thus has a **data collection** purpose.

This granular categorisation of the forms of saturation by Saunders et al. (2018) gives us an indication of how we could consider using the concept of saturation when using LLMs. Clearly LLMs have no immediate role in defining the sample or in collecting data, however they assist in the analysis. Thus, what Saunders et al. (2018) call 'inductive thematic saturation' (ITS) may be considered a useful direction of investigation in order to bring the concept of saturation within the realm of LLMs TA. The full definition of ITS that they provide is as follows, ITS: "focuses on the identification of new codes or themes and is based on the number of such codes or themes rather than the completeness of existing theoretical categories". In other words, ITS does seem to measure how codes saturate and this relates directly to the overall number of codes produced when doing QC.

Our proposition is to look at ITS to assess if LLMs also produce such saturation, or at least something looking like a form of ITS. In simple terms, we would expect that the codes produced by an LLM start to repeat across the analysis of a dataset of interviews, up to the point in which in the analysis of later interviews few or no new codes are identified by the LLM. If proved that LLMs can produce ITS in inductive initial QC we could use this as one potential metric to assess the validity of the LLM's overall inductive Thematic Analysis. We could suggest this as one metric for 'transactional validity' (TV) of LLMs' TA, where TV is an approach that "assumes that qualitative research can be more credible as long as certain techniques, methods, and/or strategies are employed during the conduct of the inquiry." (Cho and Trent, 2006, p. 322). Thus, if proven that LLMs can produce some form of ITS, then measuring LLMs ITS might be used as one of the proposed techniques to measure validity. In the context of this paper, we will apply this idea only to phase 2 of Thematic Analysis (as per Braun and Clarke, 2006 six phases model), the initial coding produced from an LLM.

In our previous proposition to perform a TA with LLMs, we presented an approach in which the LLM (GPT3.5-Turbo) produces a set/pre-defined number of codes for each interview for a given dataset (see De Paoli, 2023a). All these codes are generated independently for each interview (or chunk/portion of interview) and are then collated together in a cumulative codebook. An entirely *a posteriori* codebook reduction is then performed (again with the LLM) to identify duplicated codes. This is because each interview is indeed coded independently with the LLM having no knowledge or memory of the prior codes. To measure ITS, we propose in this paper that the reduction of duplicated codes, rather than being performed entirely *a posteriori* on the whole cumulative codebook, is instead performed after each interview. This should allow us to identify which unique new codes are created by the LLMs after each interview, and the key assumption of our work is that the absolute number of new codes given by the LLMs will decrease as the model goes through the interview's dataset. This should show therefore some form of ITS.

In the following pages, after offering a succinct review of the literature on using AI and LLMs for QA, we will describe the methodology of our experiment for attempting to measure ITS, we will present the results of the experiment including the definition of a possible metric for ITS, and we will conclude with some reflection on the use of LLMs for QA.

## Overview of the intersection between AI and QA

The intersection of QA with AI and Machine Learning (ML) is not something new in research methodologies. Researchers have sought what could be considered a data-driven approach to qualitative research, often within the broader field of Computational Social Sciences (CSS). However, compared to 'more traditional' AI approaches such as Natural Language Processing (see for example Franzosi et al., 2022) or Latent Dirichlet Allocation (see for example Evans, 2014; or Jacobs & Tschötschel, 2019), LLMs offer a potentially novel paradigm due to their capacity to manipulate and synthesize textual material, lowering the barriers for using AI for scholars with less technical skill. In the text below, we offer a brief overview of some key contributions on the intersection of QA and AI.

In an early work, Mason et al (2014) conceptualises the relation between QA and AI by highlighting how both approaches can be used in CSS and social computing. QA can provide insight into research questions and data sources, while ML and statistics contribute mathematical models and computational tools. Therefore, in this contribution more than the direct use of AI for QA, what we have is a complementarity of the approaches for CSS, with QA used to identify potential data sources and computational techniques used for data processing and analysis. Burscher et al. (2015) similarly conceptualises QA and AI as complementary. QA can provide researchers with an in-depth understanding of the social context, and AI can support researchers with larger analysis and generalisation of research results. Hence, for the authors, combining both approaches can lead to more robust research. More recently, Christou (2023) also discusses QA and AI as complementary approaches in CSS research, recognising the power of AI to aid human researchers' oversight of the analysis. Conversely, Papakyriakopoulos et al. (2021) advocates for the adoption of QA in the ML development process. QA could play a role in supporting things such as data selection, variable identification, and model architecture. QA integrated into AI development could foster more fairness and a participatory design approach in AI development. Edelman et al. (2020) conceptualise the relation between QA and AI by highlighting how CCS applies computational methods to novel sources of digital data to develop theories of human behavior. It discusses the integration of social science with computer science and emphasises the fundamental role of sociological theory for CSS. Glinka and Müller-Birn (2023) explored the human-AI collaboration in art historical image retrieval, emphasising AI's role in facilitating critical reflection in the humanities. Longo (2019) explores the impact of artificial intelligence on qualitative research methods in education and discusses how AI can be used to enhance teaching and learning. The paper suggests that AI can empower qualitative research methods by providing tools for data analysis or knowledge representation, among others. These are just some relevant examples of the intersections between AI and QA.

Several authors have consequently proposed specific tools and methods, especially for performing QA, which focus on initial coding and adopting a deductive approach. For example, Rietz et al. (2020) introduce "Cody," an AI solution for qualitative coding whose goal is to offer support to researchers to refine coding rules and perform manual annotations. This, in the authors' intentions, should bring better efficiency of the QC process and support coders inter-reliability. Chew et al. (2023) propose the novel concept of Large Language Model-assisted Content Analysis (LACA) which supports deductive coding with LLMs. In LACA, the codebook co-development and the coding are therefore driven by LLMs. The authors also emphasise (like most of the research on AI and QA) that LACA can support a reduction of the time and effort for performing a deductive CQ. In a similar vein, Gao et al. (2023) introduces "CollabCoder" which includes features such as LLM-generated code suggestions to better support collaborative CQ. Similarly, Gao et al. (2023) introduces "CoAIcoder," as an AI-assisted coding tool which, similarly to CollabCoder, is meant to improve collaboration between human researchers in QA. CoAIcoder also offers coding suggestions and supports the identification of 'coding conflicts' between researchers.

The development of new tools is therefore one of the main directions of the intersection between AI and QA. However, other contributions have focused on better understanding some of the implications for theory and methods. Several works, including those by Chen et al., 2018, and Gilles et al. (2018) have investigated the challenges of applying AI to QC. They emphasise the complexities associated with identifying ambiguity in qualitative data. Rettberg (2022) poses an interesting angle on the relation between QA and AI by proposing the use of algorithmic failure as a method to identify interesting cases for qualitative research. It argues that failed predictions in AI can reveal power dynamics and ambiguous situations that are of interest for QA. Also, others have focused on social theory's role in the intersection between QA and AI. One work highlights the importance of considering sociocultural aspects when applying machine learning in qualitative analysis (Edelman et al., 2020). Radford and Joseph (2020) emphasised the potential limits of AI in QA and stressed that these limits can be countered by social theory.

A few studies also have posed problems related to the "transactional validity" of LLM-supported qualitative data analysis. Gauthier et al. (2023) sought to conceptualise the relation between QA and ML tools by exploring how computational techniques are integrated by researchers in practice. Their work highlights the influence that these tools may exercise on the research process. This is important because it helps us see that the use of AI tools is not a neutral process, and it affects research and analysis. Consequently, Shen et al. (2023) highlight the capabilities of Large Language Models (LLMs) in analysing textual data. However, the authors also raise concerns regarding validity, privacy, and ethics. The paper emphasizes the need for researchers to validate the use of LLMs. The authors suggest the development of norms for using LLMs in social research. For example, in previous research on using LLMs for QC several authors have measured how well an LLM performs by assessing the inter-reliability with human coders using a metric know as Cohen's K (Gao et al., 2023; Xiao et al., 2023; Schiavone et al., 2023). This is an important metric to assess how well an LLM performs QC, but it has limits in its application. Inter-coder reliability can be applied largely to deductive coding, or in situations where we know with precision that the coders have seen the same material in the same order. Its application is not suitable for inductive coding, when large texts (e.g. interviews) are analysed.

In summary, the integration of ML and LLMs with QA has introduced new possibilities for research methodologies, but it also raises many issues and questions. This paper contributes to the debate about shaping ways to measure the validity of LLMs use in QA.

## Proposed Methodology

While there is an initial body of work on using LLMs for QA, we still are in a phase where we need to better assess if what we can perform with LLMs is 'good enough'. As we proposed in the introduction, exploring if some form of ITS is reached in Phase 2 of a TA performed by LLMs might constitute a useful step to assess or even just define the 'transactional validity' of the analysis. To verify this working hypothesis the research design we propose is as follows.

In our previous work (see De Paoli, 2023a) we performed phases 2 to 5 of a TA using two existing datasets of open access interviews. We have shown that an LLM (GPT3.5-Turbo) could produce most themes that human researchers (who originally collected and analysed the datasets) have themselves produced. Whilst there is not a perfect match, most of the LLM themes were deemed 'good enough' and quite similar to the one produced by the human researchers. Afterall, two human researchers independently performing a TA on a dataset of interviews may not come up with exactly the same themes anyway – some themes will overlap, and others will be different.

For this paper we propose to reuse one of these two datasets and an additional dataset of interviews to assess ITS. We will reuse the dataset related to 'teaching data sciences at UCSB' (Curty et al., 2023), which is composed of 10 interviews with university teachers/instructors of data science in a university in California. This dataset is relatively small and might not lead to significant saturation as previous research has shown that human researchers reach saturation at around 16 interviews. A second dataset that will be used is what we call the 'scrum' dataset composed of 39 interviews (Alami, 2022). This is a dataset of interviews with engineers using scrum and agile in software development. This is significantly larger than the 'teaching' dataset and should better allow to observe if there is any form of saturation.

In the work proposed in the De Paoli (2023a, further tested, refined and demonstrated in De Paoli, 2023b) the initial coding was performed as follows:

1. The model would identify – using a properly engineered and tested prompt – a pre-defined set of codes (up to a pre-defined number, which was established in order to avoid reaching the max tokens error of the LLM) from each interview. In De Paoli, 2023a we used chunks of interviews (i.e. breaking down each interview in smaller parts, or chunks) and asked the model (with an appropriate prompt) to generate up to 3-4 codes, due to the tokens limit of the model GPT3.5-Turbo (4097 tokens) that was used.
2. This would lead to a **cumulative total codebook** containing all the codes generated independently by the LLM for each interview chunk. This cumulative codebook, as such, contains repeated codes (since the model sees/codes each interview chunk separately).
3. *A posteriori*, we asked the model (again with an appropriate prompt) to operate a codebook reduction for similar/repeated codes and find therefore the **unique codes codebook**.
4. This would lead to a final codebook of unique codes, which could then be used to perform phase 3 and phase 4 of TA.

This previous workflow is briefly presented in simplified form in Figure 1.

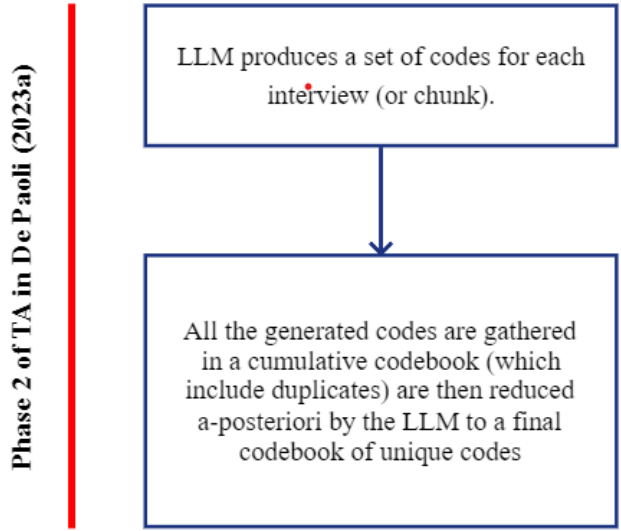

Figure 1 – General workflow for initial coding with LLMs

The process we propose to measure ITS instead will perform the check for duplicates/repeated codes after each interview rather than after all interviews have been coded by the LLM. This should allow us to identify the unique codes generated after each interview and consequently give as a precise count of the cumulative number of unique codes we can use to assess ITS. The idea is that when checking duplicates for later interviews there should be some form of ITS, that is codes will start to repeat and fewer new or unique codes will emerge. This design is as follows:

1. The model identifies a pre-defined set/number of codes from the **first** interview.
2. The model identifies a pre-defined set/number of codes from the **second** interview.
3. The codes from 1 and 2 are joined (**cumulative total codebook**) and we ask the model to identify duplicates/unique of the codes of interview two that are in the codes of interview one. If a duplicate is found, then the code already in the codebook is kept and the duplicate is discarded.
4. The unique codes from interview two are added to the codes of interview one into the **cumulative unique codebook**.
5. The model identifies a pre-defined set/number of codes from the **third** interview.
6. The model checks if the codes of interview 3 are repeated/unique in relation to the unique codebook. Unique codes are added to the unique codebook.
7. Steps 5 and 6 a are repeated for all the remaining interviews.
8. We will have at the end two lists of numbers: the total cumulative codes after each interview and the unique codes after each interview which we can use to assess what ITS could look like.

Note that for the first two interviews we need to bootstrap the unique cumulative codebook. That is the codes of the first interview are already unique. It is only after we analyse the second that we can add new unique codes. From the interview three, the process is then repeated. The workflow is presented in a simplified form in Figure 2.

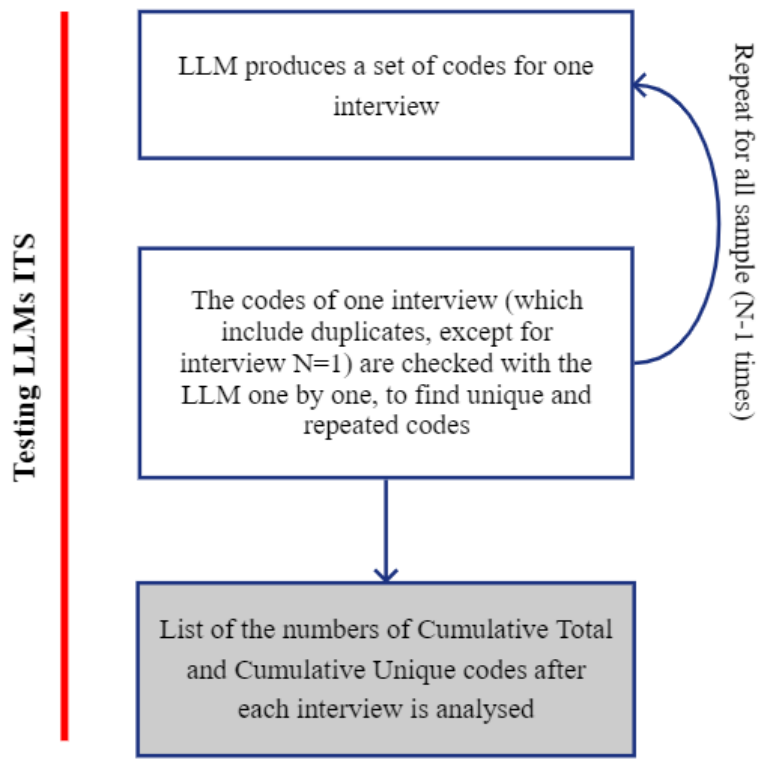

Figure 2 – Modified workflow to assessing LLMs ITS

Some additional changes compared to the previous methodology need to be further acknowledged. In De Paoli (2023a and 2023b) we worked with GPT3.5-Turbo using the OpenAI API via python scripts. GPT3.5-Turbo is limited to 4097 tokens per prompt. As we discussed there, because of this we had to divide the interviews in smaller chunks for processing and limit the number of codes generated, otherwise the model would reach the error 'max number of tokens', especially when performing the Phase 3 of TA (the generation of the Themes). Since then, a larger GPT3.5 model has been released which allows the processing of 16 thousand tokens (GPT3.5-Turbo-16k). This allows potentially better processing of the data (i.e., interviews in their entirety) and the generation of more codes per interview. This model is also relatively cheap to run compared to the newer GPT4-Turbo for example and would be suitable for an analysis even with a limited budget. Moreover, as we are not performing phase 3 of TA for this study, we do not need to avoid the 'max tokens error' happening later in the analysis. We used GPT3.5-Turbo-16k again via the OpenAI API and python. With this approach, we can programmatically pass our data (e.g. the interview texts) directly into the prompt and use python for further data manipulation or storing (e.g. in csv files).

Thus, for performing this experiment on assessing ITS we will process the interviews in their entirety (and not in chunks) and we will request the model to produce up to 15 codes for each interview for both datasets[1].

The experiment described in Figure 2 requires **two prompts**, which are passed to the model via a 'get-completion' function (please see De Paoli, 2023a for a presentation of this function), always using Temperature at 0. The first one is for the generation of the initial codes from each interview and is as follows:

```
PROMPT 1

prompt = f"""

Identify the 15 most relevant themes in the text, provide a
meaningful name for each theme in no more than 6 words, 12 words
simple description of the theme, and a max 30 words quote from the
participant.

Format the response as a json file keeping names, descriptions and
quotes together in the json, and keep them together in 'Themes'.

 ```{text}```
 """
```

This is the same prompt used in previous work (except for the increased number of codes). We should remind the readers that we use the word 'theme' in the prompt as for now it seems to work better than using 'codes' or other terms, in this kind of prompts, but for clarity this prompt generates initial codes from one interview (which is passed to the model in the {text} variable).

After each set of codes is created (i.e. after each interview) then a codebook reduction should be performed. Note that it is possible (to reduce the risk of incurring in the model timeout) to save each set of codes in separate csv files and then perform the second prompt reusing the codes from the files. For the creation of the unique cumulative codebook, we used the following prompt, which needs to run in a 'for' loop, as follows:

```
PROMPT 2

#iterates over the list of codes for one interview from prompt 1

for t in range(nr_of_codes_in_list):
```

---

[1] Note that since the counting starts from 0, requesting up to 15 means the model can generate up to 16 codes

```
    value=codes[t] #individual code t from the list of codes
    prompt = f"""
    Then, determine if value: ```{value}``` conveys the same idea
     or meaning to any element in the list cumulative_u:
    {", ".join(cumulative_u)}.
    Your response should be either a string 'true' (Same idea or
     meaning) or a string 'false' (no similarity)

     Format the response as a json file using the key
    value_in_cumulative_u
    """
```

All the codes generated from Prompt_1 for one interview (or loaded from a csv file) are included in a list 'codes' in the following format: 'code - description'. Each code ({value} in the prompt), is checked against the existing codebook {", ".join(cumulative_u)}. cumulative_u is in fact the **unique codebook**. If the code ({value}) is present in the unique codebook - when even if the text may differ but it conveys pretty much the same meaning - the model will return the string 'true'. If the model returns 'true' then it means the code already is in the codebook, otherwise the model will return the string 'false' signaling that the code is unique. After having processed all the codes, then the unique codes will be added to the cumulative_u list (the codebook of unique codes), which will then be used in the subsequent iteration (i.e. when assessing if any of the codes generated by the LLM for the next interview convey the same meaning). In this way the unique codebook will be updated after each iteration. The Prompt_2 is very similar to the one proposed in previous research (De Paoli, 2023a, 2023b) for reducing the codebook *a posteriori,* the main difference is that the response is in this case a Boolean string rather than a reduced codebook per se. This prompt is engineered in this way to make it easier to count unique codes (it is simply a matter of counting the number of 'false'), and also retrieve unique codes from the JSON generated with prompt one (or from a csv file if the JSON was stored there) to then be included in the unique codebook.

In summary, with these two prompts it is possible to obtain the following elements, both essential for assessing ITS:

1. From Prompt_1 only: A precise number of the total cumulative codes and the total cumulative codebook.

2. From Prompt_1+Prompt_2: A precise number of the codes which do not convey the same meaning for each interview in a dataset (i.e., the unique codes generated by the model for each interview).

## Results

The previous workflow and prompts were used to process the 'teaching' and 'scrum' datasets we briefly introduced earlier. We will start with the former which is composed of a smaller number of observations (n=10 interviews). The total number of codes generated with prompt 1 is 135. As we know, since the model sees each interview one at a time with no memory of the previous interview analysis and codes, the codes can be repeated. After operating prompt 2 (after each interview analysis/initial coding) the codebook results in 53 unique initial codes.

| Workflow | Number of codes |
|---|---|
| Cumulative codes **total** – prompt 1 only | 135 |
| Cumulative **unique** codes – prompt 1 + prompt 2 | 53 |

Table 1- Total and unique codebooks, 'teaching' dataset

To better understand the relationship between these codebooks we can plot the observations. First, we can look at the cumulative total codes in Figure 3.

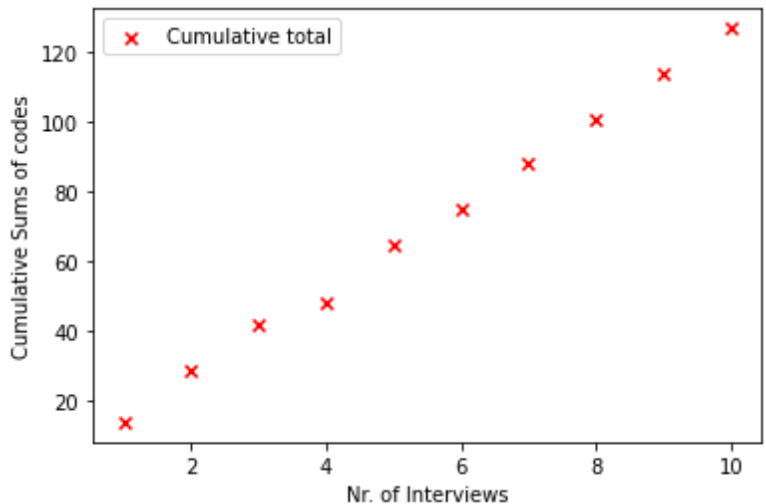

Figure 3 – Total cumulative codes scatterplot, 'teaching' dataset.

There is clearly a linear relationship as it should be expected since Prompt_1 asks to generate up to a fixed number of codes (note that for interview 4 the model generated a smaller number of codes: LLM outputs are probabilistic and even if we fix the number of codes the model may produce more or fewer). We can now plot the cumulative unique codes, as per Figure 4. We can see again a linear relation. This is perhaps less expected as one might expect this curve to look more like a 'decreasing growth curve', flattening as more interviews are coded, reflecting code saturation. Later a potential explanation for this will be offered.

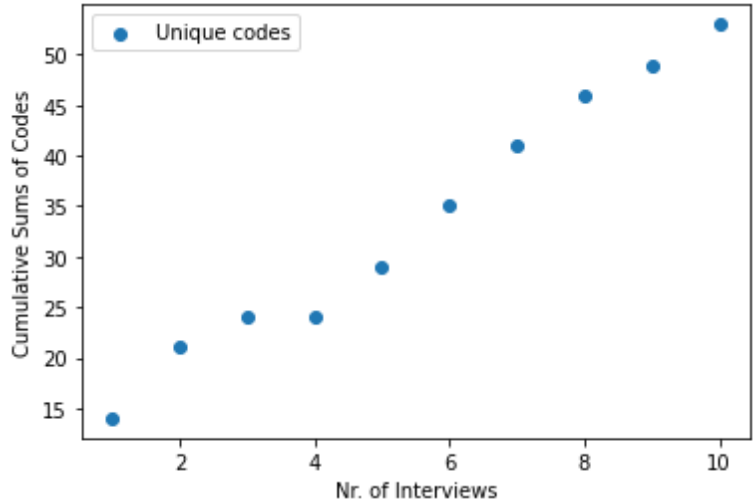

Figure 4 – Cumulative Unique codes scatterplot, 'teaching' dataset

To appreciate that there is saturation in place, we can look at both observations in the same set of axes (Figure 5). The graph in Figure 5 also renders the observations continuous by connecting the observations in the plot (but just for facilitating the interpretation, since the values are discrete). It is certainly possible also to derive a regression curve, if this was relevant. Let's first interpret the graph in Figure 5 descriptively.

The first observation is that the unique cumulative codebook grows much slower than the total codebook. A second observation is that perhaps the linear progress of the reduced codebook may be dependent on the relatively low number of interviews. Although the debate about how many interviews are needed to reach saturation is still open (which would relate to saturation as a sampling approach). Studies we mentioned in the introduction would place this figure at 12-13 (e.g. Hennik and Kaiser, 2022) or 16 (Hagman et al. 2017).

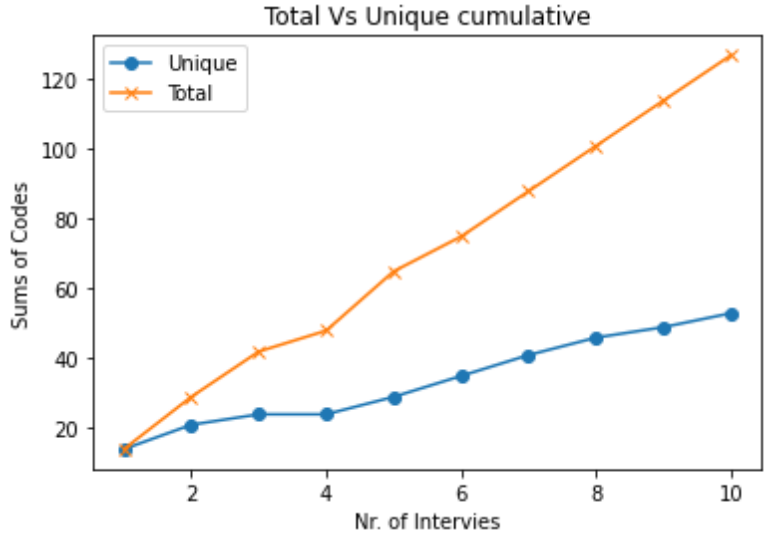

Figure 5 – Total and Unique codes comparison for the 'teaching' dataset

To evaluate the problem of whether the number of interviews have some impact on the assessment of saturation, would require producing the previous analysis with a much larger dataset. The 'scrum' dataset we propose to use is composed of 39 interviews, thus greatly exceeding the number of interviews suggested by the previous cited research and is also above one of the generic (almost word-of-mouth/hearsay) sample size figures that we hear about for qualitative research (e.g. n=20).

For the 'scrum' dataset using Prompt_1 and Prompt_2 we have obtained the number of codes in Table 2.

| Workflow | Number of codes |
| --- | --- |
| Cumulative codes **total** – Prompt_1 only | 534 |
| Cumulative **unique** codes – Prompt_1 + Prompt_2 | 66 |

Table 1- Total and unique codebooks, 'scrum' dataset

If we produce a scatter plot of the unique codes (Figure 6), we can observe in some part of the graph some potential form of saturation (with the curve flattening). However, intuitively, it looks like there still is some form of linear growth between the number of interviews and the unique codebook. The increased number of interviews thus does not seem to lead to potential saturation in the form of decreasing growth, even if this number greatly exceeds previous estimates of number required for saturation (e.g. 16).

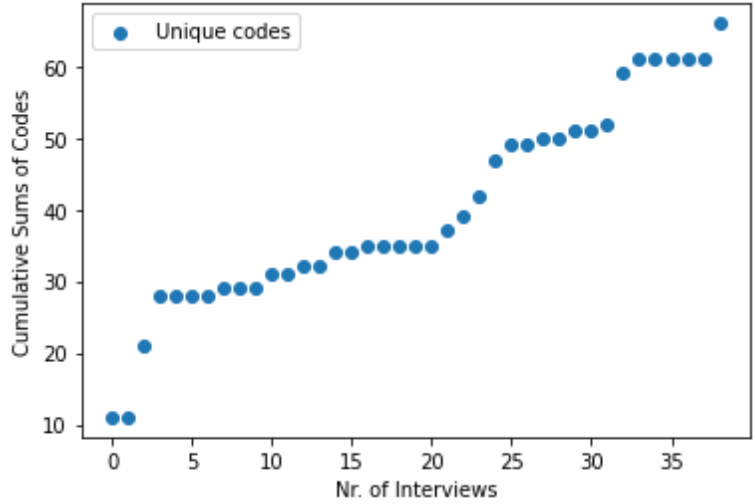

Figure 6 – Cumulative Unique codes scatterplot, 'scrum dataset'

We can plot both cumulative codebooks in the same set of axes to derive some further descriptive observations as seen in Figure 7. Again, we see in both cases the linear relations and that the cumulative total growth is much faster than the unique codebook.

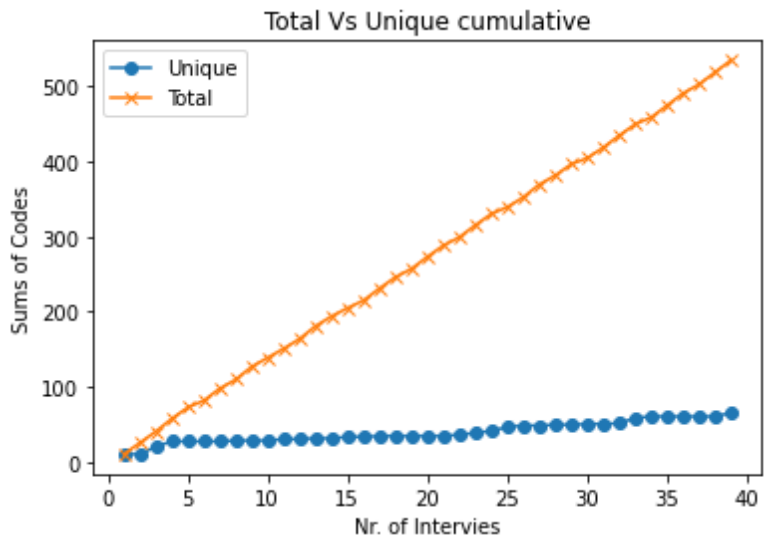

Figure 7 – Total and Unique codes comparison for the 'scrum' dataset

## Proposition for considering ITS

The linear progression of the unique codebook may be explained by the fact that the model sees each interview independently and even if the unique codebook we use to compare to each interview grows, the probability that a new code may emerge from the next interview is always greater than zero ($p>0$), due to this independence but also due to the potential size of the 'codespace' (i.e. the number of all potential valid codes). Whilst this is an interesting problem, we will defer considering it briefly in a later section of the paper. We instead concentrate now on proposing that even with a linear growth of unique codes there is still a form of saturation.

The suggestion would be to look at the differences between the observations. For example, if we look at the two lines of Figures 5 and 7 (Total/Unique), we can see how the codebook of cumulative total codes and the unique codebook increasingly diverge as the number of interviews coded increases. To an extent this divergence may signal some form of saturation. One could say that the slope of the unique codebook line is much slower than the slope of the cumulative total, and that this could amount to an inductive thematic saturation.

We could compute for each interview/observation the ratio between the number of cumulative unique codes and the number of cumulative total codes. With this we would get a measure of the rate of change between the two curves. For the 'scrum' dataset it will look like Figure 8, and intuitively for interview 1 there is no rate of change since at beginning the cumulative and reduced codebook are the same (and the ratio between the two is 1, e.g. 11 codes in both cases), however as the two curves diverge more and more the cumulative rate of change decreases (with just one exception) almost tending to a value approximatively around 0.1. This we argue is a form of saturation, it is indeed the potential equivalent of ITS for LLMs QC.

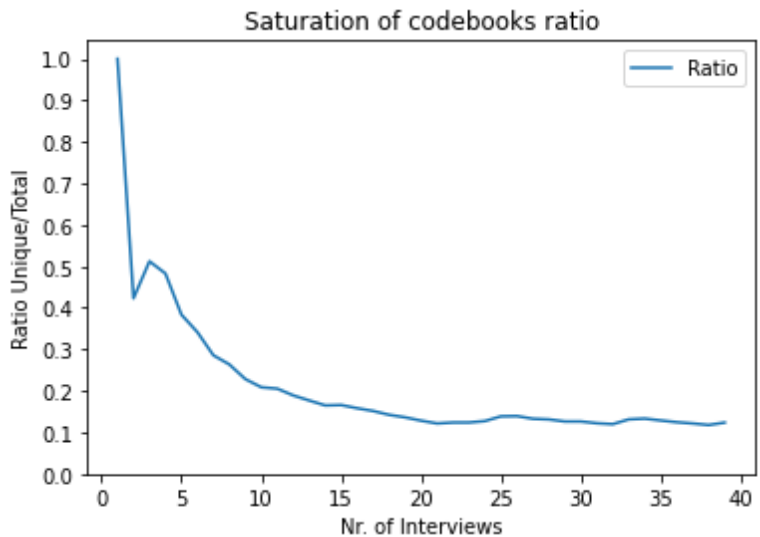

Figure 8 – ITS for the 'scrum' dataset

From this graph one may also obtain a generic indication of the number of interviews required for the slope to start to flatten. Descriptively, already around interview 10 the slope is much slower than previous interviews. At interview 20 the slope almost flattens with relatively low change.

We can do the same for the 'teaching' dataset (Figure 9) which has fewer interviews. This graph suggests saturation is starting to show, but the curve tends toward 0.4 for later interviews. That is, the steady-state ratio of the slopes is higher in this case.

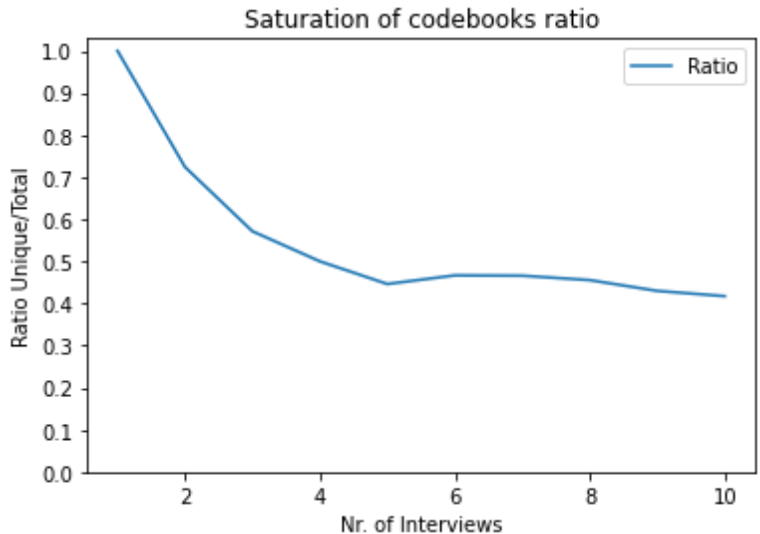

Figure 9 – ITS for the 'teaching' dataset

We infer from the shape of these curves that, even at the same 10 interview point, the scrum dataset delivers better ITS than the teacher dataset when analysed by the LLM. Therefore, we need to consider that ITS for an LLM depends also on the underlying dataset (for example if each interview in a set is more structured, the content more homogeneous, and so on). The metric consequently can also tell us something about the underlying data and may for example be useful when analysing from diverse datasets, even those collected by other researchers (e.g. Open Data taken from repositories such as zenodo).

## Relation of slopes

We would suggest that computing **the relationship between the two sets of observations --** i.e. the two codebooks (Unique/Total) -- might provide us with a synthetic indicator of how well the LLM saturates the codes, producing ITS, and also an indicator telling us something about the characteristics of the dataset itself. To help quantify this, we can use the rise over-run formula for the slope relation in the following conceptual model.

Intuitively, if Prompt_2 produces no codebook reduction at all (that is, each set of interview codes is completely different and unique form one another, for example if all the interviews came from entirely different research, conducted in different contexts, with different questions, by different researchers, etc.) then the slopes of Total and Unique will be the same, or almost the same. Conversely, we can imagine a codebook reduction process with slope at almost zero, as if we are analysing the same interview multiple times, over and over, with Temperature of the LLM at zero and obtaining always the same set of codes for each with prompt 1. In this case, there are not unique codes across the whole dataset. These are of course two extremes which should not happen in reality. But using this observation we can conclude that the relation between the slopes (assuming the linearity of the unique codebook) is a number between 0 and 1.

Therefore, the ratio between the slopes is a number between 0 and 1, the closer the ratio is to 1 the less ITS we have and vice versa, as follows:

$$0 < ITS\ (Slope\ Ratio) < 1$$

This ratio can be calculated as follows, using rise and run:

$$ITS\ (SlopeRatio) = \frac{UniqueCodes}{TotalCodes}$$

In the case of the scrum dataset this is as follows:

$$ITS(SlopeRatio) = \frac{66}{534} = 0.12$$

In the case of the 'teaching' codebooks this is as follows:

$$ITS(SlopeRatio) = \frac{53}{135} = 0.39$$

One might conclude that ITS is potentially present in both datasets, but clearly it is much higher and more significative for the 'scrum' dataset. This might depend on the underlying data (not just the number of interviews, but also how well they are structured), and not on how the model performs with the QC.

**Issue of the probability**

As discussed earlier, when describing the graphs, we observed there is a probability that unique codes are still generated after each interview even after several interviews in a potentially large sample. In fact, the unique codebook seems to progress as a linear relation with the number of interviews (at least in the two datasets we used). This is possibly dependent on the fact that each interview is analysed independently by the LLM, so each code for a new interview is an independent observation. It may also be possible that we observe a linear relation with the limited number of observations (interviews) we have, but that with a much larger number of observations (more interviews) the curve could show a decreasing growth. Let's use some statistical thinking and also the 'scrum' dataset to offer a simplified interpretation for this general problem. We will propose our interpretation of the probability problem as a thought experiment, more than as a fully worked solution, to assess the issue of the probability that at least unique code is generated when we analyse a new interview.

Let's imagine we add a new interview to the 'scrum' dataset called 'interview N+1' and we have a human analyst generate 15 codes for this interview, without knowledge of the existing unique codebook. We can ask what is the probability that at least one of these 15 codes of interview N+1 will be unique. We can calculate the Probability that at least one code from interview N+1 is unique as follows:

$$P(AtLeastOneUnique) = 1 - P(NotUnique)$$

Earlier we noted that the 'scrum' codebook has 66 unique codes. Therefore, the problem becomes that of calculating the P(NotUnique) for the additional interview, considering the existing 66 unique codes and the 15 newly generated codes by our analyst.

The 'uniqueness' of the codes clearly depends on the space of all possible codes. This number is, however, not known. We could imagine that there is potentially a large space of valid unique codes for a dataset. By no means infinite, but potentially in the order of hundreds. If we are modelling the problem like a dependent probability problem – like a bag of marbles all of different colours that we are drawing out without replacing - the probabilities of each draw are different. But, if the number of marbles is large, then the probabilities do not change all that much and might not even look like they change at all for small numbers. The graphs in Figure 10 set up a probability space of sequential numbers, then draws 15 at random. Those 15 are added to 'total' and if they are not already in 'uniques', added to it.

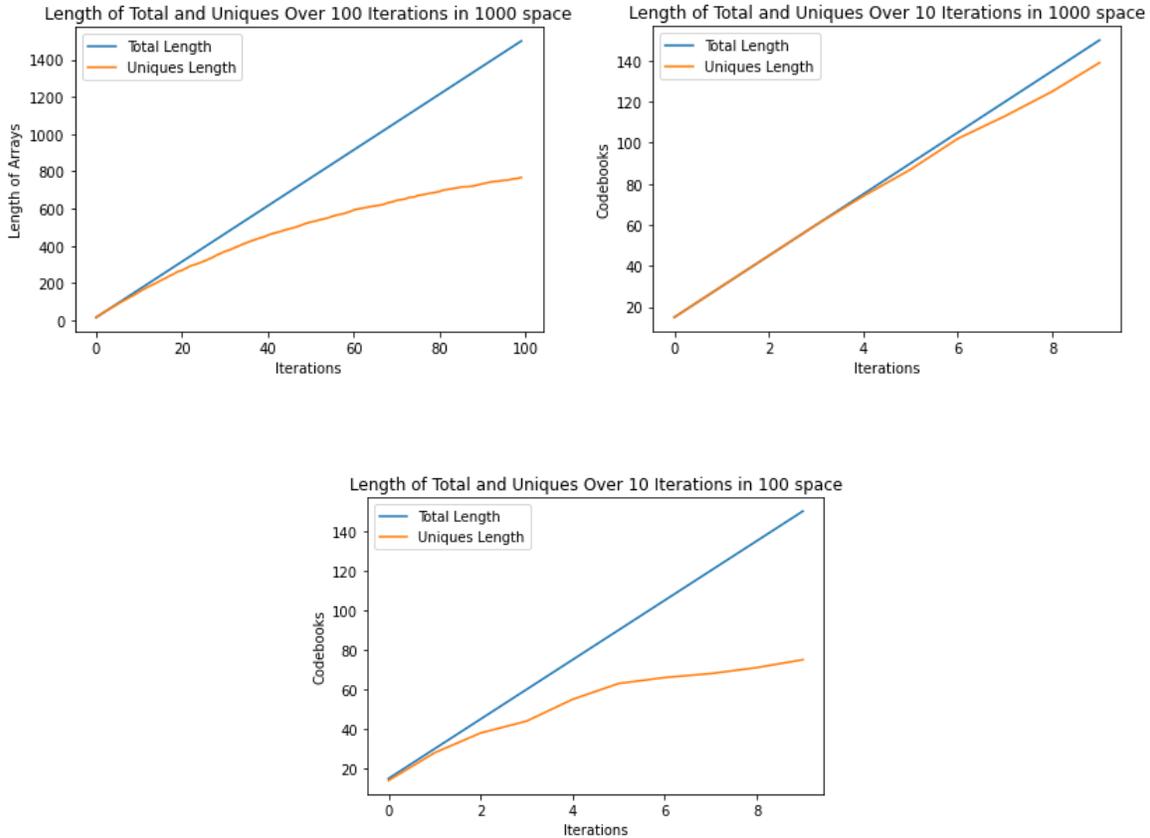

Figure 10 – Iterations in different code spaces

When the number of iterations is scaled closer to the size of the code space (e.g., 10 iterations in 100 code space or 100 iterations in 1000 code space), we see the decayed separation between total and unique we had hypothesised. But, when the size of the code space is much bigger than the number of iterations (e.g., 10 iterations in 1000 space), this is not appreciable, and they both look roughly linear (see middle graph in Figure 10). Therefore, we suppose that we are simply seeing two linear functions (in the case of unique codes for the 'teaching' and the 'scrum' datasets) because they present a relatively few iterations (i.e. interviews) in potentially large probability space (e.g. 39 iterations/interviews in a space of hundreds (or many more) valid unique codes).

From our previous probability equation, we could then model the probability function around this space as follows:

$$P(AtLeastOneUnique) = 1 - \left(\frac{UniqueCodes}{Space}\right)^{CodesInterviewN+1}$$

In the example we are proposing the UniqueCodes are 66 and the exponential (the codes for the additional interview) is 15. CodesSpace is the variable of the function, as shown in Figure 11.

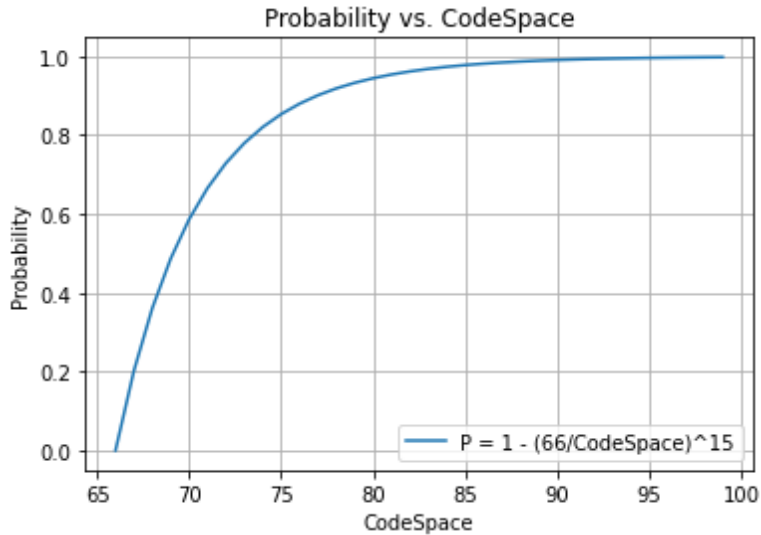

Figure 11 – Probability and CodeSpace

In fact, this function shows that the P(AtLeastOneUnique) (with Nr.CodesN+1=15) is always greater than zero, even with a relatively small CodeSpace and this P is 1 at around a CodeSpace of 90. We can interpret this as follows: the probability that a new code is generated for a N+1 interview is always greater than zero and potentially very high (approaching 1). This therefore explains potentially why we see a linear relation when building the unique codebook.

## Cosine similarity of the codes

As a last test to check that the reduction of the codebook from the total cumulative to the unique cumulative is meaningful, we compute cosine similarity (using the 66 codes from the 'scrum' analysis) with the sentence transformer all-mpnet-base-v2 (https://www.sbert.net/ - see Reimers and Gurevych, 2019). This is done to ensure that no duplicated unique codes were included in the codebook by the LLM due to e.g. potential hallucination. The figure shows that the unique codes are indeed unique. Cosine similarity value of 1 is present only in the diagonal (when similarity of a code is checked with itself, in dark red). No other cosine similarity shows the maximum value of one (no other dark red cells except in the diagonal, even though some cells show some degree of similarity with the diagonal).

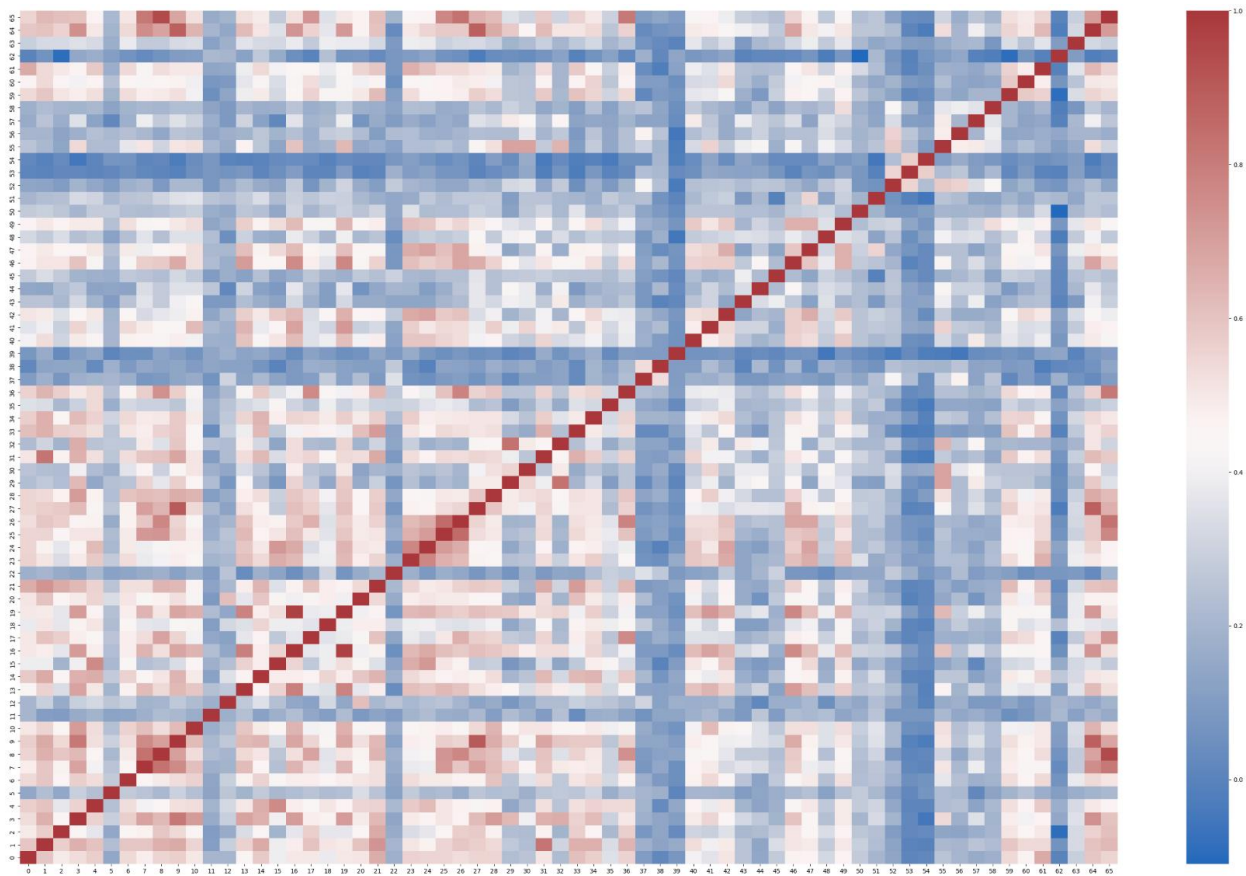

Figure 12 – Cosine similarity for the 'scrum' unique codes

Whilst this does not provide an indication of saturation per se, it shows that potentially no error or significant hallucination occurred in building the unique codebook and we can be further confident that the numerical metrics presented before can be valid.

## Discussion

The problem of this paper was to establish whether an initial coding of semi-structured interviews (phase 2 of a Thematic Analysis) performed inductively with an LLM brings about some saturation. We concentrated on the analytical properties of saturation, defined by Saunders et al. (2018) as inductive thematic saturation (ITS), which relates to how codes saturatenumerically. We postulated that ITS could be a potential approach to evaluate the 'transactional validity' of a TA performed with LLMs, that is a metric for assessing if the TA produced is of good quality. Our work also aligns with the postulation by Shen et al. (2023) about the need to imagine ways to validate and benchmark the use of LLMs for TA. Our proposed ITS metric could be one component to use for validation and benchmarking the LLM TA.

This paper has offered directions to investigate LLMs ITS (in this case GPT3.5-Turbo-16k). This will not be the final set of observations on this approach, but this manuscript offers some initial working hypotheses and empirical evidence. From the two datasets used in this work it appears that the codebook of unique codes generated by LLMs progresses largely as a linear relationship with the interviews, rather than showing a decreasing growth. Even though we postulated that with a potentially large dataset, in a relatively small code space, this may show decreasing growth. However, practically in qualitative research the number of observations (i.e. interviews) will always be a relatively small number. Nonetheless, we can still measure ITS by relating the slopes of the unique cumulative codebook and the total cumulative codebook (which is known to progress as linear relation when we request the model to generate up to n codes).

We observed that this ratio of the slopes (derived with rise and run, or more simply with the ratio of the total and unique codebooks) is a number between 0 and 1, and when the ratio starts to move away from the value of 1, we can start to appreciate that some form of ITS is reached. It is important to note that the analytical (ITS) and data sampling saturations proposed by Saunders et al. (2018) are two faces of the same coin. If LLMs do not reach ITS this may also be strongly dependent on the dataset being analysed and not just on the model capacity to saturate the codes. Moreover, the ITS metric we proposed also tells us something about the underlying dataset and not just about the LLM ITS. One can imagine that semi-structured interviews which present some degree of heterogeneity (even when coming from the same research) will lead to a lower ITS than semi-structured interviews where the structured component of the interviewing is rather solid, thus leading to a more compact unique codebook. For example, the 'scrum' dataset shows better ITS than the 'teaching' dataset even after an equivalent number of interviews is considered (i.e. 10).

Overall, the final message of this manuscript is that assessing whether the ITS ratio metric proposed in this work is a good instrument for transactional validity (Cho and Trent, 2006) will depend on further testing and verification. However, what is proposed here could be a first step in that direction.

**Implications and future work**

It is certainly up to the research community to evaluate whether the ITS metric proposed here delivered 'transactional validity' and if it can constitute a valid metric. Further testing will be required, for example with additional datasets, and possibly even with a much larger dataset than the 'scrum' dataset we used. The metric may be confirmed or disproved, but nonetheless it provides, as it is, a working hypothesis.

We are also aware that there may be an effect on saturation related to the order in which interviews are coded and consequently the order in which the codebook is reduced. This effect, in fact, has been already noted when the analysis is performed by humans. For example, Costantinou et al.

(2017) showed that changing the order in which interviews are analysed changes the point (the interview) at which saturation is reached. It will be important in future work therefore to check this effect, and possibly replicate the study by Costantinou et al. (2017) with the support of an LLM.

The approximated metric validity is also dependent on the performance in producing the unique codebook with an entirely *a posteriori* approach or with performing this after each interview. It will be important as part of future work to assess the differences between these two processes. The consideration is, however, practical more than scientific. There is a lower cost in terms of tokens processing with the *a posteriori* reduction, and there is less risk that the LLM produces a time-out error. Arguably with cheaper LLMs or Open-Source models (which nonetheless require computation), the cost aspect may become less important, and thus we can easily adopt the approach that calculates precisely the ITS metric with either the slopes or the areas of the actual lines.

Further metrics and approaches to assess the transactional validity of the LLMs QA may be explored and adopted, including different ways to measure the coding saturation. What does seem relevant is to take steps to imagine and test instruments for validity. This will allow the scientific community to assess whether QA performed with LLMs is 'good enough'

## Conclusion

This paper investigated a metric for the transactional validity of LLMs QC. It proposed a metric of analytical saturation of the codes, derived from empirical observations. Two datasets were used to perform Qualitative Coding and then assess saturation. A saturation metric was proposed in the form of the ration between the slopes of the Unique and Cumulative codebooks generated by the LLM.